\documentclass[twoside]{article}

\usepackage[accepted]{aistats2022}

\usepackage[sorting=none]{biblatex}
\usepackage{graphicx}
\usepackage{tabularx}
\usepackage{amsfonts}       
\usepackage{placeins}
\usepackage{booktabs}       
\usepackage{multirow}
\usepackage{xcolor}

\begin{document}

\twocolumn[

\aistatstitle{Improving robustness and calibration in ensembles with diversity regularization}

\aistatsauthor{ Hendrik Alexander Mehrtens \And Camila Gonz\'{a}lez \And  Anirban Mukhopadhyay }

\aistatsaddress{ Technische Universität Darmstadt} ]

\section*{\centering Abstract}

Calibration and uncertainty estimation are crucial topics in high-risk environments. We introduce a new diversity regularizer for classification tasks that uses out-of-distribution samples and increases the overall accuracy, calibration and out-of-distribution detection capabilities of ensembles. Following the recent interest in the diversity of ensembles, we systematically evaluate the viability of explicitly regularizing ensemble diversity to improve calibration on in-distribution data as well as under dataset shift. We demonstrate that diversity regularization is highly beneficial in architectures, where weights are partially shared between the individual members and even allows to use fewer ensemble members to reach the same level of robustness. Experiments on CIFAR-10, CIFAR-100, and SVHN show that regularizing diversity can have a significant impact on calibration and robustness, as well as out-of-distribution detection.

\section{Introduction}
When a machine learning system is used in high-risk environments, such as medicine and autonomous driving,  a well-calibrated estimate of the uncertainty is necessary. 
A model is said to be \textit{calibrated} \cite{CalibrationInDL} if the confidence of its predictions reflects its true probability of being correct.
However, deep neural networks tend to be overconfident in their predictions \cite{CalibrationInDL} leading to multiple recent approaches attempting to improve their calibration \cite{Pitfalls, CaliUnderDatashift}. Furthermore, models need to be robust to shifts in the data domain, which can for example arise in the data shift between the training and deployment domains.

To this day, Deep Ensembles \cite{EnsembleCalibration} outperform most other approaches. A common explanation for the improved performance is the high diversity of solutions in the ensemble \cite{BrownDiversity, DLLandscape, LiuNegCorr}, which is mostly generated by training from different parameter initializations. While this approach works well empirically,  distance in parameter space generated through training from different starting positions does not guarantee diversity in the solution space, which we refer to as \textit{functional diversity} \cite{WilsonGeneralization}. However, ensuring a diverse set of solutions in an ensemble is critical to it's performance \cite{DLLandscape, WilsonGeneralization}.

Following recent interest in the topic of diversity in neural network ensembles \cite{DLLandscape}, many publications try to implicitly generate diversity by training with different architectures \cite{DepthUncertainty, EnsembleDivSearch}, different data augmentations \cite{DivAug} and different hyperparameters \cite{HyperparaEnsembles}. However, \textbf{this approach to generate diversity is sub-optimal, as it does not guarantee diversity}. Additionally, choosing the right architectures and hyperparameters requires a lot of design decisions and is thereby time-consuming. On the other side, functional diversity can be regularized explicitly \cite{LiuNegCorr}, an idea recently used to improve adversarial robustness in ensembles \cite{DivOrthoGradient, ADP}. Although these explicit approaches guarantee diversity of predictions, they rely on diversity measures on the original training data, which can lead to a degradation in accuracy.

Additionally, these approaches \textbf{do not perform well in tasks of out-of-distribution detection} and the naive implementation requires the simultaneous training of multiple ensemble members, \textbf{which is expensive and can be prohibitive in some tasks}.

In our experiments, we put a special focus on ensembles that share parameters between the members. While these architectures require much less computational time, the lower ratio of independent parameters per member leads to a \textbf{reduction of diverse predictions} \cite{TreeNet}, which naturally lends itself to using explicit diversity maximization. For this, we use ensemble architectures with an increasing ratio of shared parameters between members and show that the effect of diversity regularization on robustness and calibration \textbf{increases with a higher ratio of shared parameters}.

We introduce the \textbf{Sample Diversity regularizer (SD)} that instead of using in-distribution images to diversify the predictions, \textbf{uses out-of-distribution images} and \textbf{ increases accuracy and calibration under dataset shift, while also increasing the out-of-distribution detection capabilities} of the model, contrary to our other baseline regularizers. The proposed regularizer can also be combined for greater effect with the other explicit diversity regularizers. Taking inspiration from the methods of Shui et al. \cite{DiversityRegularICLR}, we systematically evaluate the effectiveness of explicit diversity regularization, coming to the conclusion that \textbf{diversity regularization is especially useful when encountering dataset shift \cite{CaliUnderDatashift}, even reducing the number of ensemble members needed for the same performance} and allowing for the training of light-weight approximate ensemble architectures instead of full ensembles.  

To summarize, \textbf{our contributions} are as follows:
\begin{itemize}
    \item We introduce the \textbf{Sample Diversity regularizer}, which \textbf{increases the accuracy and calibration under dataset shift, as well as the out-of-distribution detection capabilities} and can be combined with existing diversity regularizers for greater effect.
    \item We demonstrate that diversity regularization \textbf{is highly effective for architectures with a high ratio of shared parameters}, reducing the number of needed ensemble members under dataset shift and allowing for smaller architectures.
    
\end{itemize}

 \section{Related work}
  In recent years, calibration of deep neural networks has become a focus in machine learning research. Although multiple approaches, from temperature scaling \cite{CalibrationInDL}, MC Dropout \cite{BayesDropout, BayesDropout2} to Variational Inference methods \cite{BayesByBackprop, SWAG} have been explored, 
 neural network ensembles have demonstrated that they produce the best-calibrated uncertainty estimates \cite{Pitfalls, CaliUnderDatashift, EnsembleCalibration}.
 
 An important property of well-calibrated models is whether they still give reasonable uncertainties when encountering dataset shift, as this setting better reflects real-world conditions.  Ovadia et al. \cite{CaliUnderDatashift} compared multiple approaches using the CIFAR-10-C, CIFAR-100-C and ImageNet-C datasets by Hendrycks et al. \cite{hendrycksCorruptions}, coming to the conclusion that Deep Ensembles \cite{DeepEnsemble} outperformed every other approach, making them the de-facto standard for uncertainty estimation and robustness. \par

The superiority of ensembles in these task has been partly attributed to the diversity between the individual members \cite{DLLandscape}.
 Ensemble diversity, in general, has long been a research topic in machine learning with many early works recognizing it as a key principle in the performance of ensembles \cite{BrownDiversity, LiuNegCorr}.
 Recently, a greater focus has been placed on improving diversity in neural network ensembles by different \textit{implicit} means, for example by providing each ensemble member with differently augmented inputs \cite{DivAug}, building ensembles out of different neural network architectures \cite{DepthUncertainty, EnsembleDivSearch, Swapout} or training ensemble members with different hyperparameters \cite{HyperparaEnsembles}. \par
 \textit{Explicit} approaches on the other hand try to maximize the diversity between ensemble members by orthogonalizing their gradients \cite{DivOrthoGradient}, decorrelating their predictions on all classes \cite{LiuNegCorr, DiversityRegularICLR} or on randomly sampled noise inputs \cite{MaxOverallDiv} or orthogonalizing only on non-correct classes \cite{ADP}.
 Another strategy is to increase diversity in the internal activations of the ensemble members \cite{DICE, SinhaDiversityBottleneck}, which forms a promising direction but requires computationally expensive adversarial setups. Finally, there are sampling-based methods that try to maximize the diversity of the sampling procedure, for example through Determinantal Point Processes \cite{DPP, DiverseDropout}. The advantage of these explicit approaches is that they can directly control the diversity in the ensemble and do not rely on decisions with indirect and often unclear consequences. \par

 As training ensembles is expensive, multiple methods have tried to reduce training costs. Snapshot-based methods \cite{FastGeometric, SnapshotEnsembles} save multiple epochs along a training trajectory, Batch Ensembles \cite{BatchEnsemble} generate individual ensemble members by addition of a per-member Rank-1 Hadamard-product and TreeNets \cite{TreeNet} approximate a Deep Ensemble by sharing the lower levels of a network between members. Furthermore, distillation approaches were proposed \cite{Hydra}
 that try to compress multiple networks into a single one. However, these approaches tend to reduce the diversity between the individual members, by either sharing parameters between them or not training them independently, leading to a reduction in accuracy and calibration. 
 
 In this work we show that diversity regularization is highly useful in parameter shared ensembles and that diversity regularization can not only help with accuracy and under dataset shift but also with out-of-distribution detection. Taking inspiration from Jain et al. \cite{MaxOverallDiv} we introduce an explicit diversity regularizer for classification that uses out-of-distribution samples, leaving the predictions on the original data intact.

\section{Methods and metrics}
 
 For our evaluation, we consider a classification task with $C$ classes. Given a data point $x\in \mathbb{R}^L$ out of a dataset with $N$ entries and its corresponding one-hot label $\hat y\in \mathbb{R}^C$, the prediction of the j-\textit{th} member of an ensemble with $M$ members is called  $f(x, \theta_j) =y_{j}$, where $\theta_j \in \mathbb{R}^P$ are the parameters of the j'th ensemble member. We refer to the mean of all predictions as $\bar{y}$.

In this section, we describe the evaluated regularization functions, architectures, and metrics as well as introduce our novel approach to diversity regularization.

\subsection{Regularizers}
\begin{figure*}
    \centering
    \includegraphics[width=.9\textwidth]{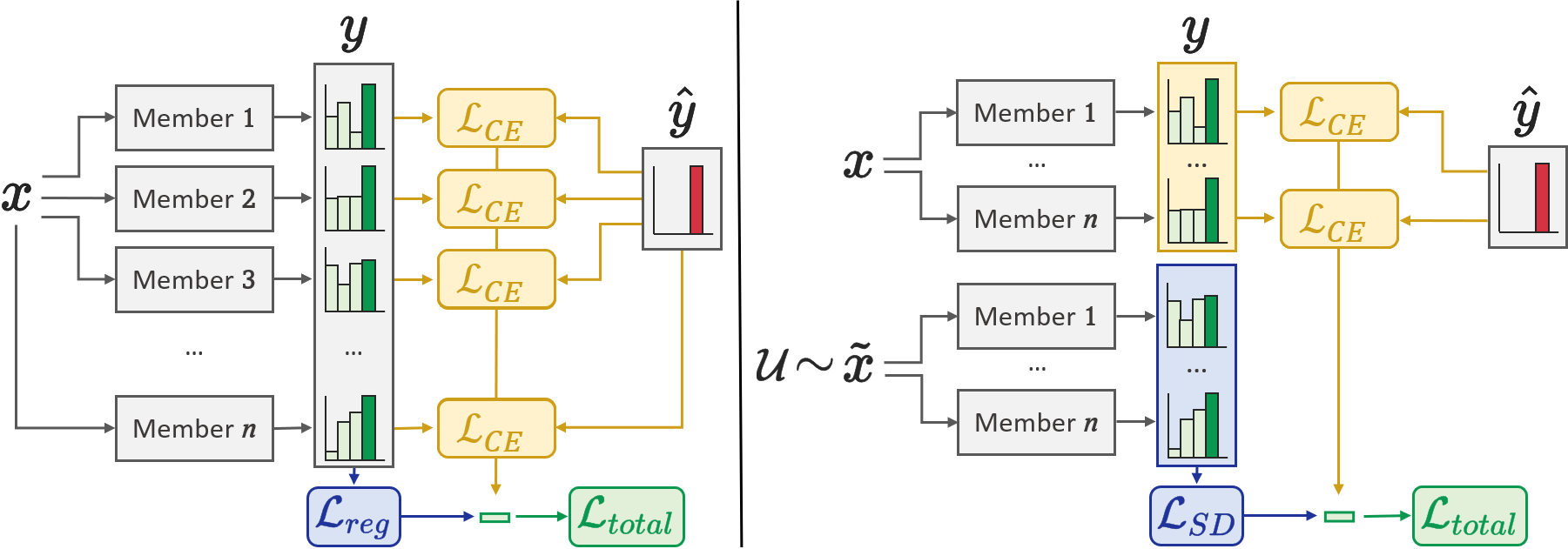}
    \caption{Conceptual figure showcasing the overall approach. Given an input \textit{x}, the individual members are not only optimized individually with regard to the cross-entropy loss but the predictions are additionally regularized by a diversity regularizer. (left) The predictions of the individual members on the original input are compared by a diversity regularizer. (right) Our \textit{Sample Diversity} (SD) approach utilizes additional uniformly sampled inputs to compute a measure of diversity as a regularizer. This preserves the original predictions on the training data.}
    \label{fig:Schema}
\end{figure*}

 Given an image $x$, a label $\hat y$ and the ensemble predictions $y_i, i \in [1,...,M]$, all regularizers $\mathcal{L}_{reg}$ work as a regularizer to the cross-entropy ($CE$) loss, where $\lambda_{reg}$ is a hyper-parameter that is chosen for each individual method.
\begin{equation}
    \mathcal{L}_{total}(\hat y,y_1,...,y_M) = \mathcal{L}_{CE}(\hat y, \bar y) - \lambda_{reg} \mathcal{L}_{reg}(\dots)
\end{equation}
For our experiments, we select a set of regularization functions that compute a measure of similarity of the individual ensemble members' predictions. An illustration of the general structure can be seen in Figure~\ref{fig:Schema}.

Regularizers under consideration are our \textbf{Sample Diversity} regularizer, the \textbf{ADP} \cite{ADP} regularizer, which was recently introduced for increasing robustness in ensembles to adversarial attacks, and the \textbf{Negative Correlation} regularizer.

Additionally we consider the average pair-wise $\chi^2$ distance (see Eq. \ref{Chi2}). All these regularizers encourage the individual members to have diverse predictions given an input and can therefore be seen as increasing the functional diversity. The regularizers will be now described:

\textbf{Negative Correlation:}
The \textit{Negative Correlation} regularizer was first used by Liu et al. \cite{LiuNegCorr} to increase the diversity in neural network ensembles. The key insight was that the error of an ensemble depends upon the correlation of the errors between individual members \cite{BrownDiversity}. Originally designed for regression tasks, it was already used by Shui et al. \cite{DiversityRegularICLR} to improve the diversity and calibration in neural network ensembles in classification tasks. This approach however reduces the accuracy of the ensemble and can easily lead to training instabilities.

\begin{equation} \label{NegCorr}
    NegCorr(y_1,..., y_M) = - \sum_{i}^{C}( (y_i - \bar{y}) \cdot (\sum_{i \neq j} y_j - \bar{y}))
\end{equation}

\textbf{ADP:} The \textit{ADP} regularizer \cite{ADP} orthogonalizes the predictions of the ensemble members on the non-correct classes during training. 

Given a correct class $k$, the vector of the predictions for the non-correct classes are formed $y_{i}^{\setminus k} = (y^{1}_i,\dots,y^{k-1}_i,y^{k+1},\dots,y^{C}_i)$, re-normalized and stacked into a matrix $Y_{\setminus k} \in \mathbb{R}^{(C-1) \times M}$. Furthermore an entropy regularizer ($H$) is used preventing extreme solutions. Together the regularizer is optimized using the hyperparameters $\alpha$ and $\beta$.
\begin{equation}
    ADP(\bar y^{\setminus k}, y^{\setminus k}_1,\dots,y^{\setminus k}_C) =  \alpha \cdot H(\bar y^{\setminus k}) + \beta \log(det(Y_{\setminus k}^T \cdot Y_{\setminus k}))
\end{equation}

\textbf{$\chi^2$ distance}:
As a distance measure between distributions, we implement the average pair-wise $\chi^2$ distance between the members' predictive distributions as a regularizer. Like the likelihood, the measure lives on the range $[0,1]$ and the regularizer can be computed as
\begin{equation} \label{Chi2}
    \chi^2(y_1,...,y_M) = \log \left(\frac{1}{M \cdot (M-1)} \sum_{i\neq j}  \sum_{k=1}^{C} \frac{y_i^{(k)} -  y_j^{(k)}}{y_i^{(k)} + y_j^{(k)}}\right)
\end{equation}

\textbf{Sample Diversity:}\label{SampleDiversity}
Building on the work of Jain et al. \cite{MaxOverallDiv} and \textit{ADP} \cite{ADP}, we introduce a similar regularizer for classification tasks, which we illustrate in Figure \ref{fig:Schema}.
Instead of regularizing diversity on the predictions of in-distribution data points, which could degrade performance, we generate out-of-distribution data points and enforce predictive orthogonality there. The loss reaches a minimum if all predictions are orthogonal on the sampled data points and thereby diverse but correct on the in-distribution data. Image batches are sampled from the uniform probability distribution and our regularizer encourages all ensemble members to have pairwise orthogonal logits on them. 
Given all logits outputs for a sampled data point $\tilde x$, of all $M$ ensemble members ($\tilde y_1, ..., \tilde y_M$), normalized to length one and stacked in a matrix $\tilde Y \in \mathbb{R}^{C \times M}$, we maximize Eq. \ref{eq:SD} as our regularizer. 

\begin{equation} \label{eq:SD}
    SampleDiversity(\tilde y_1, ..., \tilde y_M) = log(det(\tilde Y^T \cdot \tilde Y)) 
\end{equation}

In the out-of-distribution detection literature, multiple other approaches that utilize OOD data during training exist, however these approaches act on single neural networks, utilize adversarial generators and experiment in the out-of-distribution detection domain \cite{malinin2019reverse, lee2018training, hendrycks2019deep}. Our goal is to formulate a practical functional diversity regularizer that utilizes the strength of ensembles for robustness and calibration.

\subsection{Architectures}

As more shared parameters reduce the computational resources required when training an ensemble but also the diversity of the ensemble,
we wish to study if higher dependency between members, increases the viability of diversity regularization. To this end, we compare the independently trained Deep Ensembles of randomly initialized neural networks \cite{EnsembleCalibration} without adversarial training with TreeNets \cite{TreeNet} that approximate a Deep Ensemble by sharing a base part of the network with each member, as well as Batch Ensembles \cite{BatchEnsemble} that generate their members by adding a Rank-1 Hadamard product to the parameter matrices of a base network and have the least number of independent parameters. We limit the scope of our study to the aforementioned architectures, although other architectures like the MiMo architecture \cite{MiMo} exist, as they are closest in structure to the Deep Ensemble.

\subsection{Metrics}

When working with calibration it is not only important to be well-calibrated on the original data but also under reasonable dataset shifts, which is crucial for real-world application. To evaluate this, corrupted datasets are used, that simulate realistic noise and corruptions settings. All our metrics will be reported on the original datasets, as well as under dataset shift. 
Additionally to accuracy and negative log-likelihood (\textit{NLL}), we measure additional metrics, which are explained in the following:

\textbf{Calibration:} A commonly used measure of calibration is the \textit{Expected Calibration Error} ($ECE$) \cite{ECE}. As noticed by Ashuka et al. \cite{Pitfalls} this metric may not produce consistent rankings between models. For this reason, temperature scaling \cite{CalibrationInDL} with five-fold cross-validation on the test-set is deployed to generate consistent results. The temperature is computed for each dataset and intensity level separately, as temperature scaling on the original data does not guarantee being correctly scaled on the shifted data \cite{CaliUnderDatashift}. Our scores are computed after applying temperature scaling to the predictions. The temperature is chosen to minimize the negative log-likelihood, as proposed by Guo et al. \cite{CalibrationInDL}. 

\textbf{AUC-ROC}: The ability of detecting out-of-distribution data is tested, as intuitively more diverse ensemble members should produce more diverse predictions when evaluated on out-of-distribution data.
We  use the confidence of the average prediction of the ensemble as threshold classifier for distinguishing between in-distribution ($ID$) and out-of-distribution ($OOD$) data. Following \cite{Pitfalls} the AUC-ROC metric is reported for $OOD$ detection.

\begin{table*}
  \caption[Experiments on CIFAR-10]{Experiments on CIFAR-10. Comparison of diversity regularization on different architectures with ensemble size 5 under dataset shift on the original (org.) data and highest corruption level (corr.). 
  }
  \label{tab:TableC10}
  \centering
  \begin{tabularx}{0.73 \textwidth}{m{0.11 \textwidth}m{0.13\textwidth}m{0.085\textwidth}m{0.085\textwidth}m{0.085\textwidth}m{0.085\textwidth}}
    \toprule
    Model  & Method & \multicolumn{2}{c}{Accuracy $\uparrow$} & \multicolumn{2}{c}{ECE  $\downarrow$} \\
    & & org. & corr. &  org. & corr. \\ 
    
    \midrule
    
    \multirow{5}{0.12\textwidth}{DeepEns.}& ind. & $\textbf{.936}_{\pm .001 }$&$.543_{\pm .010 }$&$.023_{\pm .001 }$&$.170_{\pm .014 }$ \\ 

    & ADP & $.933_{\pm .000 }$&$.549_{\pm .005 }$&$.032_{\pm .002 }$&$\textbf{.126}_{\pm .010 }$ \\ 

    & NegCorr. & $.934_{\pm .001 }$&$.538_{\pm .002 }$&$.023_{\pm .001 }$&$.164_{\pm .007 }$ \\ 

    & $\chi^2$ & $.934_{\pm .001 }$&$.542_{\pm .006 }$&$.023_{\pm .000 }$&$.171_{\pm .008 }$ \\ 

    & \textbf{SampleDiv.} & $.933_{\pm .001 }$&$\textbf{.579}_{\pm .004 }$&$\textbf{.022}_{\pm .001 }$&$.134_{\pm .007 }$ \\ 


    \midrule
    
    \multirow{5}{0.12\textwidth}{TreeNet}& ind. & $.919_{\pm .002}$  & $.523_{\pm .01}$ &  $.035_{\pm .001}$ & $.234_{\pm .010}$ \\ 
    
    & ADP & $.917_{\pm .002}$  & $.535_{\pm .019}$ &  $\textbf{.024}_{\pm .000}$ & $\textbf{.180}_{\pm .031}$ \\ 
    
    & NegCorr. & $.918_{\pm .003 }$&$.528_{\pm .013 }$&$.027_{\pm .002 }$&$.200_{\pm .014 }$ \\ 
    
    & $\chi^2$ & $\textbf{.920}_{\pm .004 }$&$.517_{\pm .013 }$&$.027_{\pm .001 }$&$.238_{\pm .013 }$ \\ 

    & \textbf{SampleDiv.} & $.916_{\pm .002 }$&$\textbf{.545}_{\pm .007 }$&$.030_{\pm .002 }$&$.213_{\pm .014 }$ \\ 


    \midrule
    
    \multirow{5}{0.12\textwidth}{BatchEns.} & ind. & $.905_{\pm .001 }$&$.512_{\pm .019 }$&$.097_{\pm .002 }$&$.285_{\pm .014 }$ \\ 

    & ADP & $\textbf{.906}_{\pm .002 }$&$.517_{\pm .011 }$&$\textbf{.032}_{\pm .008 }$&$\textbf{.171}_{\pm .049 }$ \\ 

    & NegCorr. & $.904_{\pm .001 }$&$.503_{\pm .002 }$&$.072_{\pm .021 }$&$.258_{\pm .030 }$ \\ 

    & $\chi^2$ & $.905_{\pm .002 }$&$.503_{\pm .014 }$&$.058_{\pm .007 }$&$.265_{\pm .030 }$ \\ 

    & \textbf{SampleDiv.} & $.904_{\pm .000 }$&$\textbf{.545}_{\pm .007 }$&$.037_{\pm .015 }$&$.175_{\pm .032 }$ \\ 


    \bottomrule
    
  \end{tabularx}
\end{table*}

 \section{Experiments and results}

We first describe the general setup that is used in all of our experiments. After that, we test the effect of our different diversity regularizers on the accuracy, \textit{NLL} and calibration and later focus on  out-of-distribution detection, 
different ensemble sizes and variants of the regularizers.

\subsection{Datasets, models, and training}

The base architecture for all our experiments is a ResNet-20 \cite{ResNet}. 
We train our models on the CIFAR-10, CIFAR-100 \cite{CIFAR} and SVHN \cite{SVHN} datasets. For experiments under dataset shift, we use the corrupted versions of the CIFAR-10 and CIFAR-100 datasets created by Hendrycks et al. \cite{hendrycksCorruptions}  and additionally create a corrupted version of the SVHN dataset using all 19 corruptions with 5 levels of corruption intensity. Our experiments are implemented with PyTorch \cite{PyTorch} and run on a Nvidia Tesla T4 GPU.

 All experiments are conducted, unless otherwise stated, with a learning rate of $1e-4$, a $L_2$ weight decay of $2e-4$, a batch size of 128 and Adam \cite{Adam} as the optimizer, with the default $\beta_1$ and $\beta_2$ parameters. Each model is trained for 320 epochs. For augmentation, we use random crops and random horizontal flips, as described by Kaiming et al. \cite{ResNet}. Temperature scaling is used on each dataset and corruption intensity level individually. The optimal temperatures are computed by five-fold cross-validation on the test dataset, as suggested by Ashukha et al. \cite{Pitfalls}.
 
 When using the TreeNet architecture the ResNet is split after the second pooling operation. The cross-entropy loss is computed for each member individually and then combined. When training the Batch Ensemble, each member is trained with the same inputs at each step, so it is possible to compare the predictions of the individual members. Batch Ensemble was originally trained by splitting a batch over the ensemble members in each step. When evaluating the impact of this change, we found no significant differences between the two training methods. The comparison can be found in the supplemental material.
 
 Each experiment is performed 3 times and we report the mean performance together with the standard deviation.  Whenever possible, hyperparameters are chosen as presented in the original papers. All other parameters were fine-tuned by hand on a 10$\%$ split of the training data.
 
 When training with the \textit{ADP} regularizer, we use the parameters $\alpha=0.125$, $\beta=0.5$ which performed best for us in preliminary experiments. Those are the original parameters reported in the paper scaled by a factor of $0.25$.
 For the \textit{Sample Diversity} regularizer, we choose the number of sampled images equal to the original batch size. The images are sampled uniformly on all 3 channels in the range $[0, 1]$. We then choose $\lambda_{SD} = 0.5$ for training.
 The $\chi^2$ baseline used $\lambda_{\chi^2}=0.25$.
 The \textit{Negative Correlation} regularizer proved hard to train in a stable manner. We use $\lambda_{NC}=1e-5$, as values above this threshold destabilized the training process.
 
\subsection{Diversity regularization under dataset shift}

We train the Deep Ensemble, TreeNet, and Batch Ensemble architectures on CIFAR-10, CIFAR-100, and SVHN. The experiments are performed with 5 ensemble members. On all three datasets, we compare the independently trained ensembles, which we refer to as 'ind.' in our figures, with the regularized variants. We then evaluate all models on the corrupted versions of the datasets, comparing the accuracy, $NLL$ and $ECE$ (computed over 100 bins).

\begin{figure}
    \centering
    \includegraphics[width=1. \linewidth]{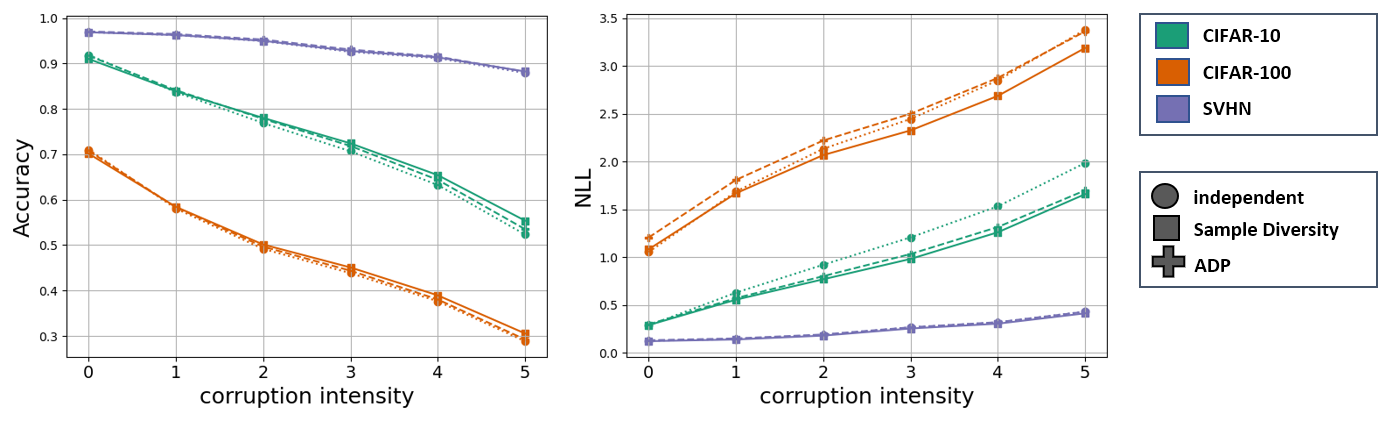}
    \caption{Accuracy (left) and \textit{NLL} (right) with different regularizers over different datasets. 
    }
    \label{fig:DatasetComp}
\end{figure}

Table \ref{tab:TableC10} shows the results of our experiments on CIFAR-10 and the corrupted variant with all architectures. We compare the accuracy and \textit{ECE} on the original data and on the highest corruption level. The results for the CIFAR-100 and SVHN datasets can be found in the supplemental material. 

The \textit{Sample Diversity} regularizer \textbf{outperforms all other regularization functions in terms of accuracy on the corrupted data}, improving the accuracy under dataset shift by $3.6\%$ (Deep Ensemble), $2.3\%$ (TreeNet) and $3.3\%$ (Batch Ensemble), as can be seen on all architectures under dataset shift.  Both the \textit{Sample Diversity} and  \textit{ADP} regularizer \textbf{outperform the other approaches in terms of ECE}. The only exception occurs on the non-corrupted data with the Deep Ensembles architecture, where the \textit{ADP} regularizer slightly decreases the calibration.
Overall the $\chi^2$ and \textit{Negative Correlation} regularizer perform worse. This is most likely due to the fact that the diversity in these regularizers is also enforced on the correct class. When training these regularizers we also observed training instabilities. 

\begin{figure}[t]
    \centering
    \includegraphics[width=1.1 \linewidth]{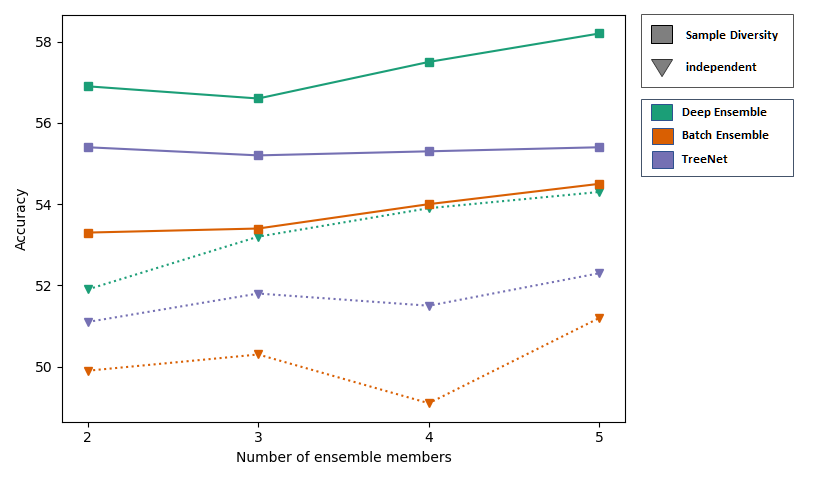}
    \caption{Comparison of the three different architectures over the ensemble sizes 2 to 5 on the highest corruption level on CIFAR-10.}
    \label{fig:EnsSizeComp}
\end{figure}

As hypothesized the \textbf{diversity regularization is also effective when using constrained ensemble architectures}. This is particularly noticeable for the Batch Ensemble architecture, which has the highest amount of shared weights per member, but also on the TreeNet architecture, a significant decrease of the \textit{ECE} is observable, even on the original data, compared to the Deep Ensemble architecture, where diversity regularization performs worse. An interesting observation is that the TreeNet and Batch Ensemble regularized with the \textit{Sample Diversity} loss \textbf{outperform the Deep Ensemble of the same size} ($54.5\%$ on both architectures, compared to $54.3\%$ for the unregularized Deep Ensemble) on the corrupted data in terms of classification accuracy.  Looking at the results it is clear that \textbf{regularizing diversity helps in improving robustness to dataset shifts}. It improves ensemble calibration, lowering the \textit{ECE} under dataset shift significantly. The displayed metrics show a clear split between \textit{ADP} and \textit{Sample Diversity} on one side and the normal training routine and 
on the other side.

Figure \ref{fig:DatasetComp} compares the mean accuracy and negative log-likelihood of the \textit{Sample Diversity} and \textit{ADP} regularizer over all three datasets. We use a TreeNet with 5 members. The x-axis denotes the corruption level, the colors encode the dataset, while the line style and marker encode the regularizer.The \textit{Sample Diversity} regularizer (solid, square) consistently improves the accuracy and decreases the negative log-likelihood under dataset shift. This difference is especially noticeable for the CIFAR-10 and CIFAR-100 datasets, on SVHN all methods stay relatively close to each other. The \textit{ADP} (dashed, plus) regularizer on the other hand can even strongly decrease the negative log-likelihood on the original data, as can be seen with the CIFAR-100 results. 

\begin{figure}[t]
     \centering
     \includegraphics[width=\linewidth]{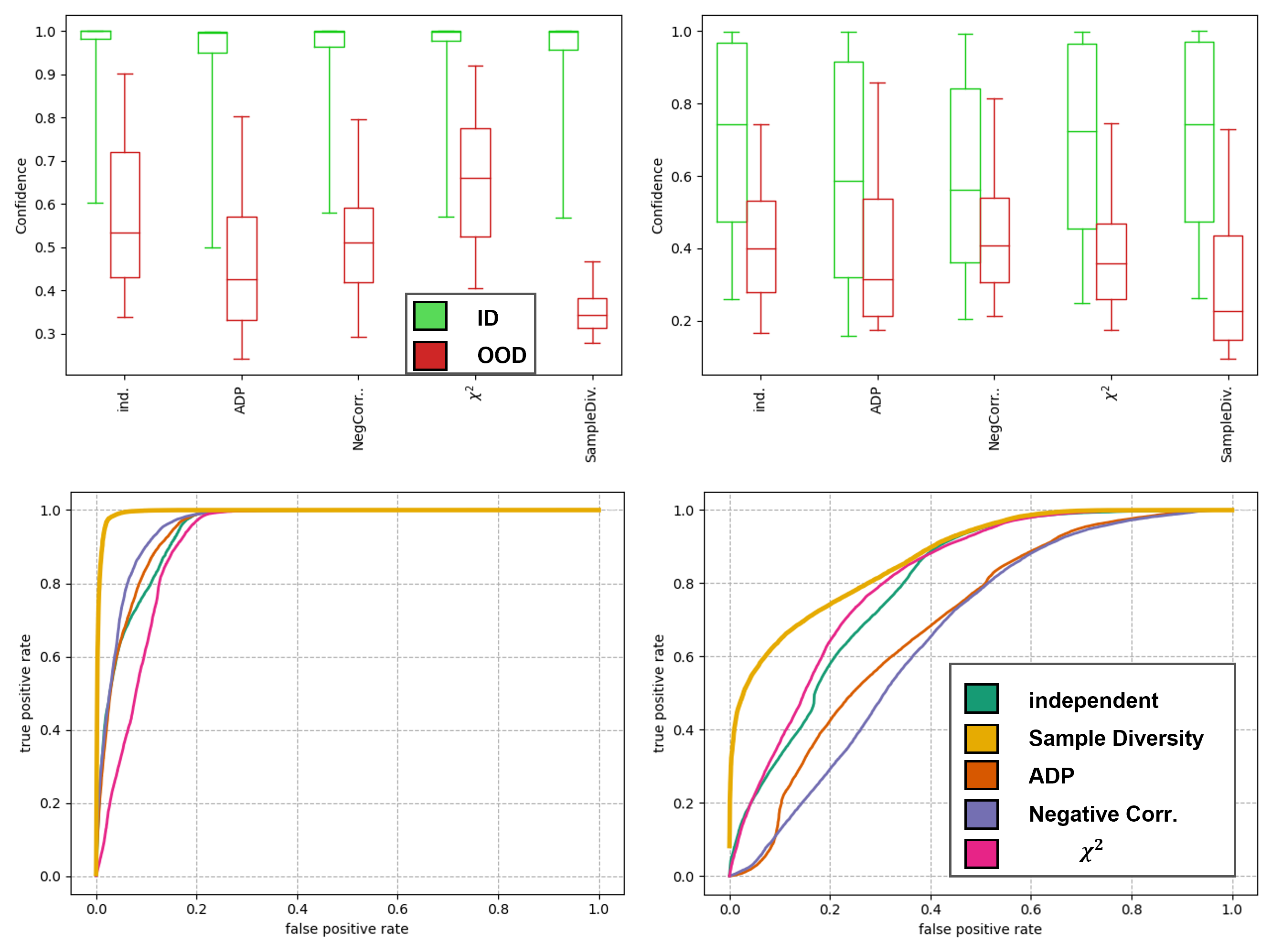}
     \caption{Distribution of confidence (top) and ROC curve for distinguishing between ImageNet and ID data (bottom) for a TreeNet architecture with 5 members with different regularization's. }
     \label{fig:MH5BoxROC} 
\end{figure}

Figure~\ref{fig:EnsSizeComp} compares the effectiveness of \textit{Sample Diversity} regularization on the highest corruption level of CIFAR-10 over the different ensemble sizes 2 to 5, comparing the mean accuracy, \textit{NLL} and \textit{ECE}. The colors encode the architecture, while the line style and marker encode the regularizer (\textit{Sample Diversity} or independent training). As can be seen in the figure, diversity regularization is even highly effective when using as few as 2 ensemble members and does not require a large pool of members. \textbf{Even a TreeNet or BatchEnsemble with 2 members, outperforms the unregularized equivalent with 5 members}. This strongly reduces the number of ensemble members required for the same performance and shows \textbf{that even lightweight ensemble architectures can outperform a Deep Ensemble}. A table with detailed results can be found in the supplemental material.

\subsection{Out-of-distribution detection}\label{OODChapter}
\begin{table}
   \caption{AUC-ROC over three runs, on separating in-distribution data and out-of-distribution data. Entries marked with '-' diverged. 
   }
   \label{TableAUCROC}
   \centering
   \begin{tabularx}{\linewidth}{lllccc}
     \toprule
     Model & Method & \multicolumn{2}{c}{AUC-ROC $\uparrow$}\\
     \multicolumn{2}{c}{(trained on)} & CIFAR-10 & CIFAR-100\\
    
     \midrule
    
      \multirow{5}{0.13\textwidth}{DeepEns.} &  indi. &$.980_{\pm .014 }$& $.798_{\pm .025 }$\\
      &  ADP &$.965_{\pm .017 }$&$.804_{\pm .034 }$\\
      &  NCL &$\textbf{.993}_{\pm .001 }$& $.729_{\pm .038 }$\\
      &  $\chi^2$ &$.983_{\pm .011 }$& $.834_{\pm .037 }$\\
      &  \textbf{SD} &$.982_{\pm .012 }$& $\textbf{.919}_{\pm .026 }$\\
     \midrule
     \multirow{5}{0.13\textwidth}{TreeNet} &  indi. &$.947_{\pm .044 }$&$.799_{\pm .049 }$\\
      &  ADP &$.952_{\pm .008 }$&$.695_{\pm .050 }$\\
      &  NCL &$.960_{\pm .011 }$&$.663_{\pm .097 }$\\
      &  $\chi^2$ &$.916_{\pm .020 }$&$.815_{\pm .019 }$\\
      &  \textbf{SD} &$\textbf{.995}_{\pm .003 }$&$\textbf{.877}_{\pm .122 }$\\
     \midrule
      \multirow{5}{0.13\textwidth}{BatchEns.} & indi. &$.928_{\pm .008 }$&$.497_{\pm .187 }$\\
       & ADP &$.909_{\pm .026 }$&$.595_{\pm .110 }$\\
       & NCL &$.934_{\pm .076 }$&-\\
       & $\chi^2$ &$.974_{\pm .008 }$& $\textbf{.809}_{\pm .122 }$\\
       & \textbf{SD} &$\textbf{.991}_{\pm .004 }$&$.614_{\pm .080 }$\\
    
     \bottomrule
    
   \end{tabularx}
\end{table}

\begin{table*}[ht!]
    \caption[Comparison of different \textit{Sample Diversity} and \textit{ADP} variants]{Comparison of different \textit{Sample Diversity} (SD) and \textit{ADP} variants on a TreeNet on CIFAR-10. }
    \label{tab:SampDivVariants}
    \centering
    \begin{tabularx}{0.73 \linewidth}{m{0.23 \linewidth}XXXXXX}
    \toprule
    Reg.  & \multicolumn{2}{c}{Accuracy $\uparrow$} & \multicolumn{2}{c}{ECE  $\downarrow$} \\
     & org. & corr. &  org. & corr. \\
    \midrule
    ADP (base) & $.917_{\pm .002}$  & $.535_{\pm .019}$ &  $\ .024_{\pm .000}$ & $.180_{\pm .031}$ \\
    SD  \ \ (base) & $.916_{\pm .002 }$&$.545_{\pm .007 }$&$.040_{\pm .002 }$&$.213_{\pm .014 }$ \\
    \midrule
    OrthoInit$_{OOD}$ & $.917_{\pm .001 }$&$.516_{\pm .007 }$&$.036_{\pm .003 }$&$.232_{\pm .010 }$ \\
    OrthoInit$_{ID+OOD}$ & $.917_{\pm .002 }$&$.512_{\pm .006 }$&$.037_{\pm .001 }$&$.247_{\pm .004 }$ \\
    \midrule
     SD (batch size 512) & $.921_{\pm .001 }$&$.562_{\pm .012 }$&$.041_{\pm .001 }$&$.210_{\pm .010 }$ \\
     SD+ADP & $.919_{\pm .001 }$&$.560_{\pm .010 }$&$.027_{\pm .002 }$&$.150_{\pm .008 }$ \\
    \midrule
    SD (adversarial) & $.918_{\pm .000 }$&$.549_{\pm .013 }$&$.036_{\pm .006 }$&$.167_{\pm .026 }$ \\
    \midrule
    SD (ImageNet) & $.916_{\pm .004 }$&$.520_{\pm .007 }$&$.040_{\pm .002 }$&$.234_{\pm .014 }$ \\
    \midrule
    SD$_{\chi^2}$  & $.919_{\pm .003 }$&$.545_{\pm .008 }$&$.041_{\pm .005 }$&$.222_{\pm .010 }$ \\
    ADP$_{\chi^2}$ & $.921_{\pm .001 }$&$.533_{\pm .004 }$&$.027_{\pm .001 }$&$.208_{\pm .007 }$ \\
    \midrule
    ind. \ \ \ \  (ensemble size 11)  & $.939_{\pm .001 }$&$.544_{\pm .005 }$&$.023_{\pm .001 }$ &$.155_{\pm .003 }$\\
    ADP$_{\chi^2}$ (ensemble size 11)  & $.935_{\pm .004 }$&$.552_{\pm .007 }$&$.028_{\pm .001 }$ &$.138_{\pm .007 }$\\
    SD$_{\chi^2}$ \ \ \  (ensemble size 11) & $.932_{\pm 0.003 }$&$.589_{\pm .003 }$&$.027_{\pm .001 }$ &$.114_{\pm .009 }$\\
    \bottomrule
    \end{tabularx}
\end{table*}

Figure \ref{fig:MH5BoxROC} shows the distribution of confidence and the receiver operating characteristic ($ROC$) for differentiating between in-distribution and out-of-distribution data, which was in this experiment chosen as tinyImageNet \cite{tinyImageNet}, a 200-class subset of the ImageNet \cite{ImageNet} dataset. The evaluated models are TreeNet architectures with 5 members, trained on CIFAR-10 and CIFAR-100. We do not report on SVHN, as every model there reached a near-perfect separation between in- and out-of-distribution data.

Looking at the ROC on CIFAR-10, \textit{Negative Correlation} and \textit{ADP} improve the separation slightly, while \textbf{Sample Diversity strongly increases the dataset separation}. Things look different on CIFAR-100 where all regularizers but \textit{Sample Diversity} \textbf{decrease the separation} of the two datasets compared to independent training. To explain this, we take a look at the differences in the confidence distributions across the datasets. While all models are fairly confident in their predictions on CIFAR-10, most likely due to the few well-separated classes, on CIFAR-100 every model is highly unconfident, with confidence distributions between in-distribution data and OOD highly overlapping.  There, the increased uncertainty that \textit{Negative Correlation} and \textit{ADP} introduce on in-distribution predictions is a disadvantage as the confidence distributions now tend to overlap more. On the other hand, \textit{Sample Diversity} that only encourages orthogonality on OOD data improves the OOD detection capability.

Table \ref{TableAUCROC} reports the AUC-ROC, as suggested by Ashuka et al. \cite{Pitfalls}, for all our evaluated models. None of the baseline regularizers is able to consistently increase the AUC-ROC over the level reached by independent training. \textit{Sample Diversity}, even though it is not the best in every single experiment, outperforms independent training and nearly all other regularizers consistently in every setting. We conclude that \textit{Sample Diversity} can \textbf{not only increase the robustness and calibration but at the same time also the out-of-distribution detection capabilities of the model, while in-distribution diversity regularization is detrimental to this task}.

\subsection{Variants and ablation studies}

Table \ref{tab:SampDivVariants} shows multiple variants of our \textit{Sample Diversity} and the \textit{ADP} regularizer. Increasing the batch size of sampled images to 512 (fourfold) \textbf{increases the accuracy on the \textbf{original} and corrupted data even further, while also lowering the \textit{ECE}}, suggesting that there is more future potential in this approach.
When \textit{Sample Diversity} is combined with the \textit{ADP} regularizer, we observe an \textbf{improvement in all measured metrics} suggesting that \textit{Sample Diversity} \textbf{combines constructively with other diversity regularizers}. We suspect that this is due to the different behaviors of the regularizers (see supplemental material) that emerge due to the different datasets the regularizers work on. Replacing the uniform noise with the tinyImageNet dataset \textbf{reduces the overall accuracy on the original and corrupted data and increases the calibration error}. We suspect the ever-new nature of newly sampled images to be more effective, than a limited pool of images, for diversity maximization on OOD data. Combining real datasets with augmentations and corruptions could be a future solution to the problem.

We replace the log-determinant regularization term in the \textit{ADP} ($ADP_{\chi^2}$) and \textit{Sample Diversity} ($SD_{\chi^2}$) regularizer, with the pair-wise $\chi^2$ distance (see Eq. \ref{Chi2}). This has the advantage that ensembles \textbf{with more members than classes can be trained with diversity regularization}, as otherwise for a matrix $Y \in \mathbb{R}^{C \times M}$ ($C-1$ in case of ADP)
\begin{equation}
    det(Y^T\cdot Y) =  0, \: if \; C < M  
\end{equation}
Table \ref{tab:SampDivVariants} shows that this formulation performs just as well while being numerically more stable, as it does not require a determinant and matrix-inversion operation and allows for arbitrary ensemble sizes. An ensemble of size 11 on CIFAR-10 in this case also \textbf{benefits greatly from diversity regularization. This is a useful property for datasets with a low number of classes.}

Finally, following recent work \cite{MaxOverallDiv, lee2018training} we apply adversarial Fast-Gradient sign attacks \cite{goodfellow2015explaining} on the regularizer, to create more effective out-of-distribution images, on which to maximize the diversity. However, this approach only slightly increases the accuracy on the corrupted data ($0.3\%$), suggesting that uniformly sampled images are already effective enough. 

To test if regularization during test-time is necessary or if a functional diverse initialization is enough, we apply only the regularizers in a 3 epoch warm-up phase. We test only using \textit{Sample Diversity} (OrthoInit$_{OOD}$) and \textit{Sample Diversity} combined with \textit{ADP} (OrthoInit$_{IID+OOD}$). These initialization do not lead to improvements, suggesting that constant regularization during training is necessary. 

\section{Conclusion}
We introduce the \textit{Sample Diversity} regularizer, which is well suited for \textbf{improving accuracy and ECE and can be combined with the \textit{ADP} regularizer for greater effect}. Contrary to other regularizers, our regularizer \textbf{also increases the out-of-distribution detection capabilities}.
Our experiments show that \textbf{diversity regularized ensembles are better in terms of accuracy and calibration under dataset shift}. Regularizing ensembles beyond the diversity reached by independent training especially on architectures with shared parameters is beneficial.
Even  \textbf{the TreeNet and Batch Ensemble can outperform a Deep Ensemble in terms of robustness to dataset shift when diversity regularization is used}, even when we use fewer members. Our final experiments indicate that the \textit{ADP} loss formulation is sub-optimal and future research could increase the viability of diversity regularization further.

\printbibliography

\newpage

\onecolumn

\newpage
\section{Supplemental Material}

\subsection{Distance in parameter space}

We additionally use the distance in parameter space as a baseline regularizer to contrast distance in parameter space with functional diversity. 
Distance in parameter space is no guarantor of diverse functions, as different parameter settings can represent the same function, through a reparamerization of the function.
A simple  example of this is a permutation of filters inside a convolutional layer, with corresponding changes to the filters in adjacent layers. While this operation creates distance in parameter space, both parameter settings represent the same function. 

Benjamin et al. \cite{benjamin2019measuring} experimented with distance in function space, coming to the conclusion that parameter distance is no good measure for functional differences. We conduct similar experiments by introducing the \textit{WeightCos} regularizer that orthogonalizes the parameter vectors $\theta_i \in \mathbb{R}^{P}$, stacked into a matrix $\Theta \in \mathbb{R}^{M \times P}$, of the individual ensemble members during training.

\begin{equation*}
    WeightCos(\theta_1,\dots, \theta_M)= log(det(\Theta \cdot \Theta^T))
\end{equation*}

We report the performance of the \textit{WeightCos} regularizers in some of the later tables. 
While \textit{WeightCos} does not lead to improvements in the Deep Ensemble and TreeNet architecture we observed that in the BatchEnsemble architecture \textit{WeightCos} behaves very similar to the \textit{ADP} regularizer in terms of \textit{ECE} and \textit{NLL}, leading to substantial improvements under dataset shift. We suspect that the formulation of the individual members in the BatchEnsemble does not allow for easy reparametrization. In this case orthogonality in parameter space could induce functional diversity, which is an intersting finding for future work.
See Section \ref{AppendixResultsDatasets} for the results including the \textit{WeightCos} regularizer.

\subsection{Differences in diversity regularizers}

To further investigate the effects diversity regularization has on the predictions of the ensemble, we measure the predictive entropy, Jensen-Shannon-Divergence, and Oracle NLL \cite{TreeNet}. The predictive entropy of the ensemble is an indicator of how diverse an ensemble is, as more varying individual predictions will increase the entropy of the mean prediction.

\begin{equation}
    H(\bar y) = -\frac{1}{\log{(C)}} \sum^{C}_{i=1} \bar y_i \log{(\bar y_i)}
\end{equation}

However, an ensemble composed of highly uncertain but similar members will also produce high entropy predictions. For this, we measure the Jensen-Shannon-Divergence (abbr.: JSD), which is the mean Kullback-Leibler (abbr.: KL) divergence between the individual ensemble members' predictions $(y_1,\dots, y_M)$ and the mean prediction $\bar{y}$. A higher value for the \textit{JSD} therefore indicates more diverse predictions across the ensemble.
\begin{gather}
        KL(y_i||y_j) = - \sum_{k=1}^{C} y_i^{(k)} \cdot log(\frac{y_i^{(k)}}{y_j^{(k)}}) \\
    JSD(\bar{y}, y_1,\dots,y_M) = \frac{1}{M} \sum^{M}_{i=1} KL(y_i||\bar{y})
\end{gather}

A measure introduced by Lee et al. \cite{TreeNet} is the Oracle NLL, which is the negative log-likelihood of the best performing ensemble member for each input. A more diversified ensemble with more specialized members results in a lower Oracle \textit{NLL}.

 \begin{figure}[t]
    \centering
    \includegraphics[width=\linewidth]{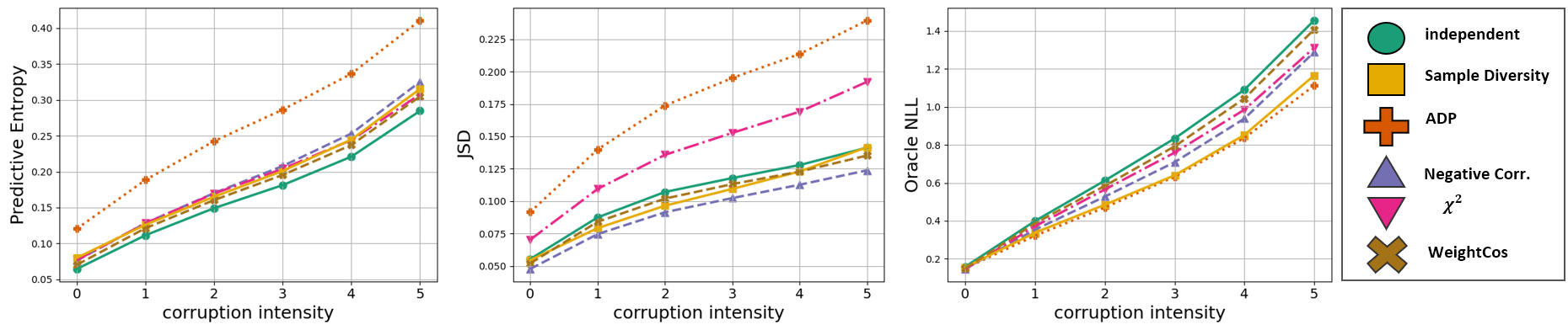}
    \caption{\textit{Entropy} (left), \textit{Jensen-Shannon-Divergence} (middle) and \textit{Oracle NLL} (right)  for a 5-member TreeNet on CIFAR-10, with different regularizations. The x-axis indicates the level of corruption.}
    \label{fig:EntropyMH5}
\end{figure}

The result for a TreeNet with 5 members can be found in Figure \ref{fig:EntropyMH5}. The markers indicate the type of regularization and the x-axis indicates the level of corruption. It can be seen that in the \textit{ADP} regularized ensemble the individual members stray the furthest from the mean prediction, as seen in the higher entropy, Jensen-Shannon-Divergence between the members and also the significantly lower \textit{Oracle NLL} \cite{TreeNet}. On the other hand, the entropy of the \textit{Sample Diversity} regularizer is only slightly increased compared to the independent ensemble training, even though the \textit{ECE} and accuracy are constantly superior, which is most likely due to \textit{Sample Diversity} only regularizing on out-of-distribution data, leaving the predictions on the training data intact. \textit{Sample Diversity} and \textit{ADP} have the lowest \textit{Oracle NLL}, indicating a high functional diversity and specialised members. Consistent with our prior results, this is then followed by the $\chi^2$ and \textit{Negative Correlation} regularizer, which also performed worse on the other measured metrics. Independent training and \textit{WeightCos} have the highest \textit{Oracle NLL}, indicating that they lack diverse members. Additionally, \textit{WeightCos} also has a very low \textit{JSD}, showcasing that distance in parameter space is no guarantor of diverse members or diverse predictions.  The \textit{JSD} of the \textit{Negative Correlation} regularizer is lower than that of the independent baseline, while the predictive entropy is higher. We interpret that as the \textit{Negative Correlation} regularizer producing highly spread out and uncertain predictive distributions, which results in a lower \textit{JSD}. These results show that all regularization approaches increase the differences between member predictions, as seen in the lower \textit{Oracle NLL} and the higher entropy, but they do not all behave in the same way.
We suspect these differences in the behaviour to be the reason why \textit{ADP} and \textit{Sample Diversity} combined so well in our experiments.

\begin{figure}[t]
    \centering
    \includegraphics[width=0.8\linewidth]{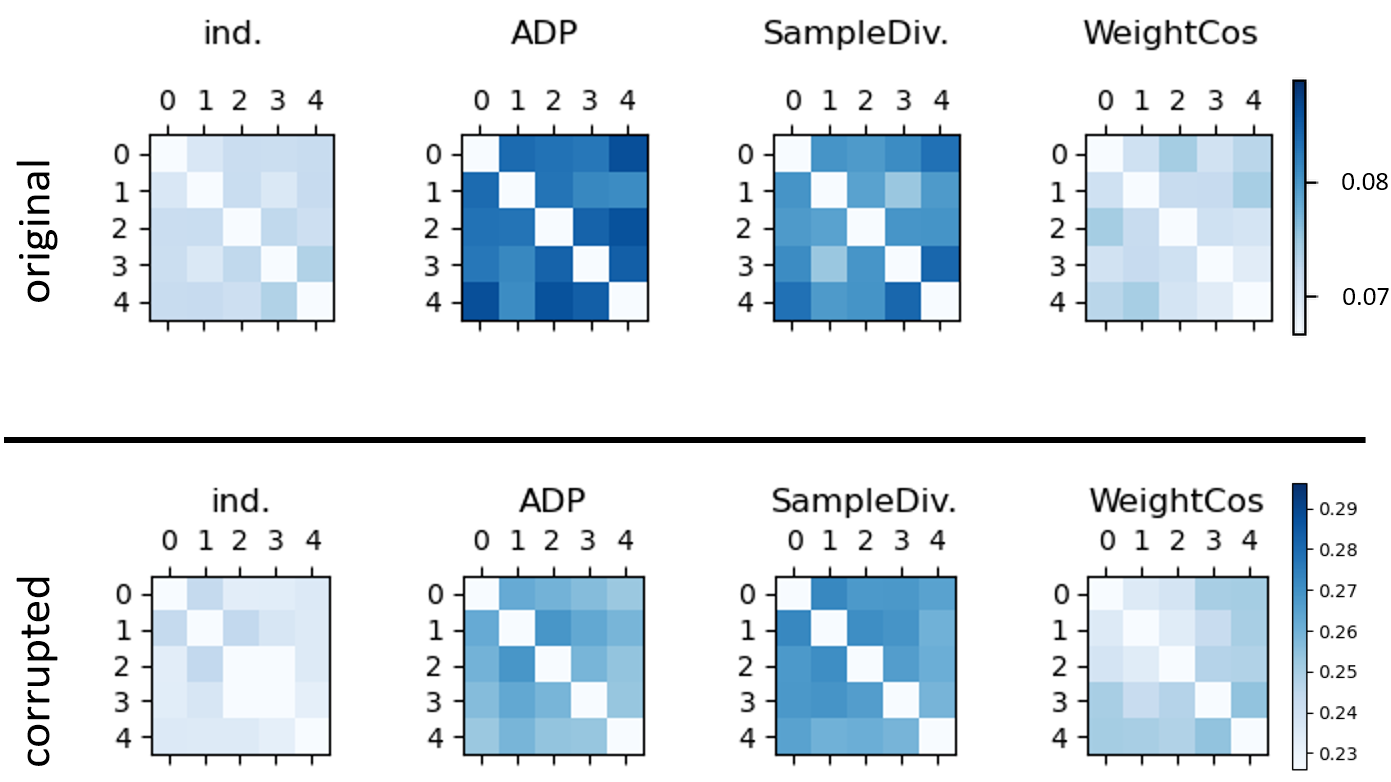}
    \caption{Percentage of different argmax predictions between the ensemble members on the original data and the highest corruption level.}
    \label{fig:PredDiffMatrix}
\end{figure}

Figure \ref{fig:PredDiffMatrix} shows the percentage of differing argmax predictions on the original data and the highest corruption level. While distance in parameter space does not lead to more diverse predictions both \textit{Sample Diversity} and \textit{ADP} produce comparably diverse predictions.

\subsection{Network capacity}

To test if diversity regularzation also works with larger models we conduct experiments on the ResNet-44 architecture. 
Table \ref{tab:Ablation} shows that when we use a bigger base architecture for the TreeNet, diversity training is still highly effective. In this case \textit{ADP} even slightly increases the accuracy on the original data. Also in terms of \textit{ECE} and \textit{NLL} both regularizers lead to an improvement. However, further experiments with more varied architectures could give more insight in future work.

\begin{table}[ht]
     \centering
     \caption{Capacity experiments. To test if, diversity regularization still performs well with bigger architectures, the backbone of the TreeNet was exchanged for a ResNet-44 architecture (Res44). All experiments were conducted on CIFAR-10 with a TreeNet architecture with 5 members. We report the accuracy, \textit{ECE} and \textit{NLL}.}
     \begin{tabularx}{\textwidth}{XXXXXXX}
     \toprule
      Method & \multicolumn{2}{c}{Accuracy $\uparrow$} & \multicolumn{2}{c}{ECE  $\downarrow$} & \multicolumn{2}{c}{NLL $\downarrow$} \\
     \multicolumn{1}{c}{(corruption intensity)} & org. & corr. &  org. & corr. & org. & corr. \\
      \midrule

     ind. & $.922_{\pm .001 }$&$.510_{\pm .016 }$&$.049_{\pm .005 }$&$.284_{\pm .004 }$&$.343_{\pm .031 }$&$2.175_{\pm .029 }$\\

    ADP & $\textbf{.926}_{\pm .002 }$&$.544_{\pm .004 }$&$\textbf{.032}_{\pm .001 }$&$\textbf{.227}_{\pm .031 }$&$\textbf{.318}_{\pm .004 }$&$\textbf{1.853}_{\pm .120 }$\\

    SampleDiv. &   $.922_{\pm .003 }$&$\textbf{.546}_{\pm .008 }$&$.047_{\pm .002 }$&$.238_{\pm .021 }$&$.321_{\pm .025 }$&$1.887_{\pm .118 }$\\

     \bottomrule
     \end{tabularx}
     \label{tab:Ablation}
 \end{table}

\subsection{Differences in training Batch Ensemble} \label{BE-Difference}
\begin{figure}[ht]
    \centering
    \includegraphics[width=\linewidth]{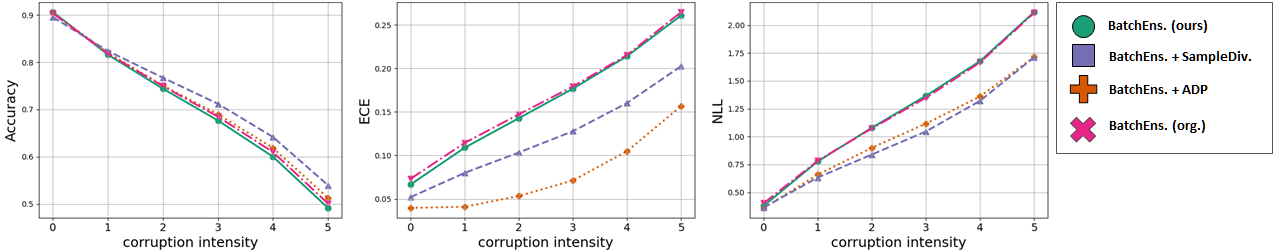}
    \caption{Comparison of training the Batch Ensemble architecture with our training schedule compared to the original implementation.}
    \label{fig:BEDiff}
\end{figure}

In our experiments, we trained the Batch Ensemble with the same data input in each step, while in the original paper \cite{BatchEnsemble} a batch was split over each member in every step. The difference is that in our approach each member sees the data points in the same order.  This could lead to a reduction in diversity, which could be larger than the gain from diversity regularization. Figure \ref{fig:BEDiff} plots our \textit{Batch Ensemble} training schedule (BatchEns. (ours)), the original training schedule (BatchEns. (org.)) and our schedule trained with \textit{Sample Diversity} and \textit{ADP}. While there is a minimal gain in the original training schedule, as noted by Ford et al. \cite{DLLandscape}, the gain of the diversity regularization is far greater.

\subsection{Different split levels}

We test using different split points, which we call split levels, in the TreeNet architecture to measure the influence of the ratio of shared parameters on the effectiveness of diversity regularization. In most of our experiments we set the split level to 3, which corresponds to a split just before the third ResNet block. A split of 0 is the same as a Deep Ensemble trained with the same data order during training. Split level 1 and 2 are placed before the first and second ResNet block, while split level 4 splits the network just before the last convolutional layer.

Table \ref{tab:DiffSplitLvls} shows our results for splitting a TreeNet trained on CIFAR-10 with 5 members at these different split points. As can be seen diversity regularization is effective even in the extreme case of split level 4, where just one convolutional layer is regularized.  

\begin{table}[ht]
  \caption[Experiments with different split levels]{Experiments with different split levels on the TreeNet architecture on CIFAR-10 with 5 members. Comparison of the \textit{Sample Diversity} and  \textit{ADP} regularizer.}
  \label{tab:DiffSplitLvls}
  \centering
  \begin{tabular}{m{0.04 \textwidth}m{0.13\textwidth}m{0.08\textwidth}m{0.08\textwidth}m{0.08\textwidth}m{0.08\textwidth}m{0.08\textwidth}m{0.08\textwidth}}
    \toprule
    Split & Method & \multicolumn{2}{c}{Accuracy $\uparrow$} & \multicolumn{2}{c}{ECE  $\downarrow$} & \multicolumn{2}{c}{NLL $\downarrow$} \\
    & & org. & corr. &  org. & corr. & org. & corr. \\
    \midrule
     & ind. & $\textbf{.936}_{\pm .001 }$&$.543_{\pm .010 }$&$\textbf{.023}_{\pm .001 }$&$.170_{\pm .014 }$&$\textbf{.210}_{\pm .004 }$&$1.783_{\pm .051 }$\\
    0 & ADP & $.933_{\pm .000 }$&$.549_{\pm .005 }$&$.032_{\pm .002 }$&$\textbf{.126}_{\pm .010 }$&$.241_{\pm .004 }$&$\textbf{1.606}_{\pm .037 }$\\
     & SampleDiv. & $.933_{\pm .001 }$&$\textbf{.573}_{\pm .004 }$&$.022_{\pm .001 }$&$.134_{\pm .002 }$&$.221_{\pm .004 }$&$1.634_{\pm .010 }$\\
    \midrule
     & ind. & $\textbf{.936}_{\pm .001 }$&$.546_{\pm .014 }$&$\textbf{.024}_{\pm .000 }$&$.163_{\pm .011 }$&$\textbf{.214}_{\pm .004 }$&$1.769_{\pm .064 }$\\
    1 & ADP & $.935_{\pm .001 }$&$.553_{\pm .005 }$&$.036_{\pm .004 }$&$\textbf{.108}_{\pm .010 }$&$.245_{\pm .008 }$&$\textbf{1.545}_{\pm .034 }$\\
     & SampleDiv. & $.933_{\pm .002 }$&$\textbf{.570}_{\pm .001 }$&$.024_{\pm .001 }$&$.148_{\pm .008 }$&$.214_{\pm .009 }$&$1.566_{\pm .029 }$\\
    \midrule
     & ind. & $\textbf{.932}_{\pm .002 }$&$.530_{\pm .002 }$&$\textbf{.023}_{\pm .001 }$&$.185_{\pm .009 }$&$\textbf{.226}_{\pm .005 }$&$1.822_{\pm .063 }$\\
    2 & ADP & $.931_{\pm .001 }$&$.539_{\pm .004 }$&$.028_{\pm .002 }$&$\textbf{.148}_{\pm .014 }$&$.256_{\pm .003 }$&$\textbf{1.671}_{\pm .046 }$\\
     & SampleDiv.  & $.929_{\pm .003 }$&$\textbf{.554}_{\pm .009 }$&$.024_{\pm .001 }$&$.166_{\pm .016 }$&$.225_{\pm .007 }$&$1.624_{\pm .074 }$\\
    \midrule
     & ind.  & $\textbf{.919}_{\pm .001 }$&$.523_{\pm .007 }$&$.035_{\pm .001 }$&$.234_{\pm .012 }$&$\textbf{.286}_{\pm .006 }$&$1.986_{\pm .076 }$\\
    3 & ADP  & $.917_{\pm .002 }$&$.535_{\pm .019 }$&$\textbf{.024}_{\pm .000 }$&$\textbf{.180}_{\pm .031 }$&$.298_{\pm .005 }$&$1.699_{\pm .132 }$\\
     & SampleDiv.  & $.916_{\pm 0.002 }$&$\textbf{.545}_{\pm .007 }$&$.030_{\pm .002 }$&$.213_{\pm .014 }$&$.305_{\pm .013 }$&$1.822_{\pm .044 }$\\
    \midrule
     & ind. & $.899_{\pm .003 }$&$.502_{\pm .010 }$&$.067_{\pm .018 }$&$.257_{\pm .032 }$&$.395_{\pm .051 }$&$1.996_{\pm .209 }$\\
    4 & ADP & $\textbf{.902}_{\pm .001 }$&$.509_{\pm .024 }$&$\textbf{.038}_{\pm .002 }$&$\textbf{.216}_{\pm .027 }$&$\textbf{.393}_{\pm .008 }$&$\textbf{1.843}_{\pm .128 }$\\
     & SampleDiv. & $.900_{\pm .001 }$&$\textbf{.527}_{\pm .007 }$&$.081_{\pm .018 }$&$.243_{\pm .022 }$&$.438_{\pm .063 }$&$1.902_{\pm .093 }$\\
    \bottomrule
  \end{tabular}
\end{table}

\FloatBarrier
\newpage
\subsection{Detailed results - CIFAR-10, CIFAR-100 and SVHN} \label{AppendixResultsDatasets}

Here we show the detailed results over all three datasets. Table \ref{tab:TableC10Detailed} shows our  results on CIFAR-10, Table \ref{tab:SVHN} our results on the SVHN dataset and Table \ref{tab:C100} our  results on CIFAR-100. We notice that all regularizers have problems on CIFAR-100, which is most likely due to the large number of classes compared to the small number of ensemble members. 

When we repeat the experiments with ensemble size 20 on CIFAR-100 we observe a better performance (see Table \ref{tab:C100Ens20}). However, datasets with large number of classes remain a problem. Here methods that utilize class number independent measures like internal activation's of the neural network for diversification could prove superior.

\FloatBarrier

\subsection{Detailed results - Different Ensemble Sizes} \label{AppendixResultsEnsembleSize}

Table \ref{tab:DiffEnsSizeTreeNet} shows the results for our experiments on CIFAR-10 on the TreeNet architecture with different ensemble sizes, displaying the accuracy, \textit{ECE} and \textit{NLL}. We compare the ensemble sizes 2 to 5, using the \textit{ADP} and \textit{Sample Diversity} regularizer. Table \ref{tab:DiffEnsSizeBatchEns} shows the same for the experiments on the BatchEnsemble and Deep Ensemble architectures. 
As before the scores are computed over 3 differently seeded runs and we report the mean and standard deviation. \textit{Sample Diversity} consistently improves the accuracy under dataset shift and also lowers the \textit{ECE} compared to the independent ensemble training (ind.). \textit{ADP} performs best in terms of calibration error, having the lowest \textit{ECE} in most settings. As mentioned before even a TreeNet of size 2 can outperform a Deep Ensemble of size 5 on the corrupted data, if diversity regularization is used. 
Furthermore, the \textit{$ADP_{\chi^2}$} and  \textit{$SampleDiv.{\chi^2}$} formulation allow for training a Deep Ensemble of size 11 on CIFAR-10, which still provides large gains in terms of robustness to dataset shift.

\begin{table}[ht]
  \caption[Experiments on CIFAR-10]{Experiments on CIFAR-10 with five members on different architectures.}
  \label{tab:TableC10Detailed}
  \centering
  \begin{tabularx}{\textwidth}{m{0.11 \textwidth}m{0.13\textwidth}m{0.085\textwidth}m{0.085\textwidth}m{0.085\textwidth}m{0.085\textwidth}m{0.085\textwidth}m{0.085\textwidth}}
    \toprule
    Model  & Method & \multicolumn{2}{c}{Accuracy $\uparrow$} & \multicolumn{2}{c}{ECE  $\downarrow$} & \multicolumn{2}{c}{NLL $\downarrow$} \\
    & & org. & corr. &  org. & corr. & org. & corr. \\
    
    \midrule
    
    \multirow{6}{0.12\textwidth}{DeepEns.}& ind. & $\textbf{.936}_{\pm .001 }$&$.543_{\pm .010 }$&$\textbf{.023}_{\pm .001 }$&$.170_{\pm .014 }$&$\textbf{.210}_{\pm .004 }$&$1.783_{\pm .051 }$\\

    & ADP & $.933_{\pm .000 }$&$.549_{\pm .005 }$&$.032_{\pm .002 }$&$\textbf{.126}_{\pm .010 }$&$.241_{\pm .004 }$&$\textbf{1.606}_{\pm .037 }$\\

    & NegCorr. & $.934_{\pm .001 }$&$.538_{\pm .002 }$&$\textbf{.023}_{\pm .001 }$&$.164_{\pm .007 }$&$\textbf{.210}_{\pm .001 }$ & $1.714_{\pm .082 }$\\

    & $\chi^2$ & $.934_{\pm .001 }$&$.542_{\pm .006 }$&$\textbf{.023}_{\pm .000 }$&$.171_{\pm .008 }$&$.226_{\pm .007 }$&$1.767_{\pm .044 }$\\

    & SampleDiv. & $.933_{\pm .001 }$&$\textbf{.579}_{\pm .004 }$&$.022_{\pm .001 }$&$.134_{\pm .007 }$&$.217_{\pm .008 }$&$1.494_{\pm .026 }$\\

    & WeightCos. & $.935_{\pm .001 }$&$.537_{\pm .007 }$&$\textbf{.023}_{\pm .000 }$&$.166_{\pm .003 }$&$.212_{\pm .004 }$&$1.790_{\pm .034 }$\\

    \midrule
    
    \multirow{6}{0.12\textwidth}{TreeNet}& ind. & $.919_{\pm .002}$  & $.523_{\pm .01}$ &  $.035_{\pm .001}$ & $.234_{\pm .010}$ &  $.286_{\pm .006}$ & $1.990_{\pm .076}$ \\
    
    & ADP & $.917_{\pm .002}$  & $.535_{\pm .019}$ &  $\textbf{.024}_{\pm .000}$ & $\textbf{.180}_{\pm .031}$&  $.298_{\pm .005}$ & $1.699_{\pm .132}$ \\
    
    & NegCorr. & $.918_{\pm .003 }$&$.528_{\pm .013 }$&$.027_{\pm .002 }$&$.200_{\pm .014 }$&$\textbf{.271}_{\pm .009 }$&$1.785_{\pm .095 }$\\
    
    & $\chi^2$ & $\textbf{.920}_{\pm .004 }$&$.517_{\pm .013 }$&$.027_{\pm .001 }$&$.238_{\pm .013 }$&$.282_{\pm .014 }$&$1.925_{\pm .117 }$\\

    & SampleDiv. & $.916_{\pm .002 }$&$\textbf{.545}_{\pm .007 }$&$.030_{\pm .002 }$&$.213_{\pm .014 }$&$.305_{\pm .013 }$&$1.822_{\pm .044 }$\\

    & WeightCos. & $.919_{\pm .002 }$&$.517_{\pm .014 }$&$.032_{\pm .003 }$&$.225_{\pm .020 }$&$.275_{\pm .014 }$&$1.927_{\pm .143 }$\\

    \midrule
    
    \multirow{6}{0.12\textwidth}{BatchEns.} & ind. & $.905_{\pm .001 }$&$.512_{\pm .019 }$&$.097_{\pm .002 }$&$.285_{\pm .014 }$&$.455_{\pm .003 }$&$2.254_{\pm .099 }$\\

    & ADP & $.906_{\pm .002 }$&$.517_{\pm .011 }$&$.032_{\pm .008 }$&$\textbf{.171}_{\pm .049 }$&$.363_{\pm .036 }$&$1.735_{\pm .160 }$\\

    & NegCorr. & $.904_{\pm .001 }$&$.503_{\pm .002 }$&$.072_{\pm .021 }$&$.258_{\pm .030 }$&$.385_{\pm .052 }$&$2.086_{\pm .230 }$\\

    & $\chi^2$ & $.905_{\pm .002 }$&$.503_{\pm .014 }$&$.058_{\pm .007 }$&$.265_{\pm .030 }$&$.391_{\pm .032 }$&$2.069_{\pm .178 }$\\

    & SampleDiv. & $.904_{\pm .000 }$&$\textbf{.545}_{\pm .007 }$&$.037_{\pm .015 }$&$.175_{\pm .032 }$&$\textbf{.343}_{\pm .014 }$&$\textbf{1.649}_{\pm .121 }$\\

    & WeightCos.  & $\textbf{.907}_{\pm .003 }$&$.499_{\pm .005 }$&$\textbf{.022}_{\pm .001 }$&$.182_{\pm .022 }$ &$.385_{\pm .005 }$&$1.836_{\pm .108 }$\\

    \bottomrule
    
  \end{tabularx}
\end{table}

\begin{table}[ht]
    \caption[Experiments on SVHN]{Experiments on SVHN with five members on different architectures.}
    \label{tab:SVHN}
    \centering
    \begin{tabularx}{\textwidth}{m{0.11 \textwidth}m{0.13\textwidth}m{0.085\textwidth}m{0.085\textwidth}m{0.085\textwidth}m{0.085\textwidth}m{0.085\textwidth}m{0.085\textwidth}}

    \toprule
    Model & Method & \multicolumn{2}{c}{Accuracy $\uparrow$} & \multicolumn{2}{c}{ECE  $\downarrow$}  & \multicolumn{2}{c}{NLL $\downarrow$} \\
    & & org. & corr. &  org. & corr. & org. & corr. \\
    
    \midrule
    
     & ind. & $.969_{\pm .001 }$&$.879_{\pm .000 }$&$\textbf{.008}_{\pm .000 }$&$.026_{\pm .006 }$ &$.126_{\pm .004 }$&$.432_{\pm .026 }$\\
 
     & ADP & $\textbf{.971}_{\pm .001 }$&$\textbf{.882}_{\pm .004 }$&$.012_{\pm .001 }$&$.009_{\pm .003 }$ &$.132_{\pm .005 }$&$.430_{\pm .014 }$\\

     TreeNet & NegCorr. & $.969_{\pm .001 }$&$.878_{\pm .004 }$&$.008_{\pm .001 }$&$.029_{\pm .009 }$ &$.126_{\pm .004 }$&$.437_{\pm .042 }$\\

     & $\chi^2$ & $.969_{\pm .000 }$&$.880_{\pm .006 }$&$.009_{\pm .002 }$&$.027_{\pm .013 }$ &$.128_{\pm .011 }$&$.434_{\pm .058 }$\\

     & SampleDiv. & $.969_{\pm .001 }$&$\textbf{.882}_{\pm .004 }$&$.009_{\pm .001 }$&$\textbf{.008}_{\pm .001 }$ &$\textbf{.122}_{\pm .004 }$&$\textbf{.413}_{\pm .018 }$\\

    \midrule
    
     \multirow{6}{0.12\textwidth}{BatchEns.} & ind. & $.965_{\pm .000 }$&$.877_{\pm .001 }$&$\textbf{.010}_{\pm .001 }$&$.028_{\pm .008 }$ &$.139_{\pm .001 }$&$.435_{\pm .008 }$\\

     & ADP & $\textbf{.970}_{\pm .001 }$&$\textbf{.891}_{\pm .004 }$&$.021_{\pm .005 }$&$.036_{\pm .018 }$ &$.139_{\pm .003 }$&$\textbf{.407}_{\pm .007 }$\\

     & NegCorr. & $.964_{\pm .003 }$&$.878_{\pm .007 }$&$\textbf{.010}_{\pm .002 }$&$.025_{\pm .008 }$ &$.138_{\pm .009 }$&$.432_{\pm .027 }$\\

     & $\chi^2$ & $.969_{\pm .002 }$&$.882_{\pm .006 }$&$\textbf{.010}_{\pm .001 }$&$\textbf{.021}_{\pm .020 }$ &$\textbf{.131}_{\pm .015 }$&$.417_{\pm .041 }$\\

     & SampleDiv. & $.965_{\pm .000 }$&$.879_{\pm .002 }$&$\textbf{.010}_{\pm .001 }$&$.029_{\pm .004 }$ &$.139_{\pm .003 }$&$.442_{\pm .013 }$\\

    & WeightCos  & $.966_{\pm .001 }$&$.867_{\pm .003 }$&$.018_{\pm .004 }$&$.022_{\pm .011 }$&$.141_{\pm .003 }$&$.450_{\pm .007 }$\\
    \bottomrule
    \end{tabularx}
\end{table}

\begin{table}[t]
    \caption[Experiments on CIFAR-100]{Experiments on CIFAR-100 with five members on different architectures.}
    \label{tab:C100}
    \centering
    \begin{tabularx}{\textwidth}{m{0.11 \textwidth}m{0.13\textwidth}m{0.085\textwidth}m{0.085\textwidth}m{0.085\textwidth}m{0.085\textwidth}m{0.085\textwidth}m{0.085\textwidth}}

    \toprule
    Model & Method & \multicolumn{2}{c}{Accuracy $\uparrow$} & \multicolumn{2}{c}{ECE  $\downarrow$} & \multicolumn{2}{c}{NLL $\downarrow$} \\
    & & org. & corr. &  org. & corr.  & org. & corr. \\
    
    \midrule
    
    \multirow{6}{0.12\textwidth}{DeepEns.} & ind. & $\textbf{.726}_{\pm .001 }$&$.300_{\pm .001 }$&$.055_{\pm .000 }$&$.058_{\pm .002 }$ &$\textbf{1.008}_{\pm .008 }$&$3.329_{\pm .008 }$\\

     & ADP & $.719_{\pm .001 }$&$\textbf{.308}_{\pm .002 }$&$.128_{\pm .004 }$&$\textbf{.032}_{\pm .004 }$ &$1.175_{\pm .003 }$&$3.274_{\pm .025 }$\\

     & NegCorr & $.623_{\pm .015 }$&$.267_{\pm .002 }$&$.073_{\pm .003 }$&$.035_{\pm .002 }$ &$1.375_{\pm .054 }$&$3.342_{\pm .016 }$\\

    & $\chi^2$  & $.717_{\pm .002 }$&$\textbf{.308}_{\pm .002 }$&$.078_{\pm .002 }$&$.035_{\pm .002 }$&$1.020_{\pm .003 }$&$3.225_{\pm .019 }$\\

     & SampleDiv. & $.708_{\pm .002 }$&$.306_{\pm .001 }$&$\textbf{.053}_{\pm .003 }$&$.053_{\pm .003 }$ &$1.070_{\pm .006 }$&$\textbf{3.173}_{\pm .025 }$\\
     
     & WeightCos  & $.726_{\pm .003 }$&$.303_{\pm .002 }$&$.056_{\pm .001 }$&$.056_{\pm .002 }$&$\textbf{1.005}_{\pm .003 }$&$3.287_{\pm .014 }$\\

    \midrule
    
     \multirow{6}{0.12\textwidth}{TreeNet} & ind. & $.710_{\pm .004 }$&$.288_{\pm .007 }$&$.036_{\pm .002 }$&$.075_{\pm .004 }$ &$\textbf{1.054}_{\pm .007 }$&$3.380_{\pm .038 }$\\

     & ADP & $.708_{\pm .004 }$&$.292_{\pm .001 }$&$.106_{\pm .004 }$&$\textbf{.020}_{\pm .002 }$ &$1.207_{\pm .006 }$&$3.362_{\pm .025 }$\\
    
     & NegCorr. & $.595_{\pm .003 }$&$.244_{\pm .008 }$&$.038_{\pm .003 }$&$.057_{\pm .008 }$ &$1.451_{\pm .016 }$&$3.474_{\pm .077 }$\\

     & $\chi^2$ & $\textbf{.711}_{\pm .003 }$&$.290_{\pm .001 }$&$.043_{\pm .002 }$&$.075_{\pm .007 }$ &$1.076_{\pm .010 }$&$3.362_{\pm .019 }$\\

     & SampleDiv. & $.701_{\pm .003 }$&$\textbf{.306}_{\pm .004 }$&$\textbf{.034}_{\pm .003 }$&$.082_{\pm .001 }$ &$1.083_{\pm .011 }$&$\textbf{3.188}_{\pm .036 }$\\

     & WeightCos & $.705_{\pm .003 }$&$.283_{\pm .001 }$&$.038_{\pm .004 }$&$.076_{\pm .005 }$ &$1.060_{\pm .005 }$&$3.437_{\pm .013 }$\\

    \midrule

     \multirow{5}{0.12\textwidth}{BatchEns.} & ind. & $.645_{\pm .000 }$&$.259_{\pm .005 }$&$.073_{\pm .005 }$&$.093_{\pm .011 }$ &$1.336_{\pm .008 }$&$3.525_{\pm .027 }$\\

     & ADP & $.637_{\pm .004 }$&$.265_{\pm .004 }$&$.082_{\pm .005 }$&$\textbf{.017}_{\pm .003 }$ &$1.537_{\pm .003 }$&$3.542_{\pm .055 }$\\ 
    
    
    & $\chi^2$  & $.642_{\pm .004 }$&$.258_{\pm .002 }$&$.055_{\pm .004 }$&$.067_{\pm .004 }$&$1.341_{\pm .019 }$&$3.484_{\pm .047 }$\\
    
    & SampleDiv. & $.635_{\pm .003 }$&$\textbf{.275}_{\pm .002 }$&$.073_{\pm .007 }$&$.102_{\pm .007 }$ &$1.372_{\pm .007 }$&$\textbf{3.354}_{\pm .002 }$\\

    & WeightCos & $\textbf{.648}_{\pm .002 }$&$.265_{\pm .003 }$&$\textbf{.038}_{\pm .003 }$&$.075_{\pm .008 }$&$\textbf{1.277}_{\pm .015 }$&$3.506_{\pm .026 }$\\

    \bottomrule

    \end{tabularx}

\end{table}

\begin{table}[ht]
    \caption[Experiments on CIFAR-100 with ensemble size 20]{Experiments on CIFAR-100 with a TreeNet and ensemble size 20. }
    \label{tab:C100Ens20}
    \centering
    \begin{tabularx}{\textwidth}{m{0.15\textwidth}m{0.1\textwidth}m{0.1\textwidth}m{0.1\textwidth}m{0.1\textwidth}m{0.1\textwidth}m{0.1\textwidth}}
    \toprule
    Method & \multicolumn{2}{c}{Accuracy $\uparrow$} & \multicolumn{2}{c}{ECE  $\downarrow$} & \multicolumn{2}{c}{NLL $\downarrow$} \\
    & org. & corr. &  org. & corr.  & org. & corr. \\

    ind. & $\textbf{.721}_{\pm .003 }$&$.297_{\pm .005 }$&$.043_{\pm .002 }$&$.083_{\pm .003 }$&$\textbf{.991}_{\pm .009 }$&$3.345_{\pm .042 }$\\
    ADP & $.718_{\pm .001 }$&$\textbf{.309}_{\pm .006 }$&$.189_{\pm .002 }$&$\textbf{.048}_{\pm .004 }$&$1.339_{\pm .013 }$&$3.319_{\pm .034 }$\\
    SampleDiv. & $.719_{\pm .002 }$&$\textbf{.309}_{\pm .007 }$&$\textbf{.042}_{\pm .001 }$&$.087_{\pm .004 }$&$1.006_{\pm .008 }$&$\textbf{3.215}_{\pm .052 }$\\
    
    \bottomrule

    \end{tabularx}

\end{table}

\begin{table}[ht]
  \caption[Experiments with different TreeNet ensemble sizes on CIFAR-10]{Experiments with different TreeNet ensemble sizes on CIFAR-10. Comparison of the \textit{Sample Diversity} and  \textit{ADP} regularizer with independent training on different architectures under dataset shift.}
  \label{tab:DiffEnsSizeTreeNet}
  \centering
   \begin{tabular}{m{0.09 \textwidth}m{0.035\textwidth}m{0.13\textwidth}m{0.08\textwidth}m{0.08\textwidth}m{0.08\textwidth}m{0.08\textwidth}m{0.08\textwidth}m{0.08\textwidth}}

    \toprule
    Model & Size & Method & \multicolumn{2}{c}{Accuracy $\uparrow$} & \multicolumn{2}{c}{ECE  $\downarrow$} & \multicolumn{2}{c}{NLL $\downarrow$} \\
    && & org. & corr. &  org. & corr. & org. & corr. \\

    \midrule
    
     \multirow{3}{0.12\textwidth}{TreeNet} & & ind.  & $\textbf{.909}_{\pm .002 }$&$.511_{\pm .005 }$&$.036_{\pm .000 }$&$.220_{\pm .013 }$ &$\textbf{.310}_{\pm .004 }$&$1.862_{\pm .026 }$\\

     &2 & ADP  & $\textbf{.909}_{\pm .000 }$&$.523_{\pm .012 }$&$\textbf{.028}_{\pm .002 }$&$\textbf{.172}_{\pm .015 }$ &$.330_{\pm .007 }$&$1.696_{\pm .069 }$\\

     &  & SampleDiv. & $.906_{\pm .002 }$&$\textbf{.541}_{\pm .009 }$&$.042_{\pm .002 }$&$.192_{\pm .007 }$ & $.332_{\pm .010 }$&$\textbf{1.680}_{\pm .032 }$\\

    \midrule
    
    \multirow{3}{0.12\textwidth}{TreeNet} &  & ind. & $\textbf{.919}_{\pm .002 }$&$.518_{\pm .008 }$&$.039_{\pm .002 }$&$.236_{\pm .003 }$ & $\textbf{.295}_{\pm .004 }$&$1.972_{\pm .012 }$\\

      & 3 & ADP & $\textbf{.919}_{\pm .002 }$&$.525_{\pm .006 }$&$\textbf{.025}_{\pm .002 }$&$.189_{\pm .009 }$ & $.301_{\pm .013 }$&$1.752_{\pm .029 }$\\
     
      &  & SampleDiv. & $.910_{\pm .001 }$&$\textbf{.542}_{\pm .014 }$&$.036_{\pm .001 }$&$\textbf{.187}_{\pm .011 }$ & $.303_{\pm .006 }$&$\textbf{1.693}_{\pm .059 }$\\

    \midrule

    \multirow{3}{0.12\textwidth}{TreeNet} &  & ind. & $.918_{\pm .002 }$&$.515_{\pm .009 }$&$.034_{\pm .003 }$&$.226_{\pm .012 }$ & $\textbf{.290}_{\pm .008 }$&$1.933_{\pm .120 }$\\

     & 4 &  ADP & $\textbf{.919}_{\pm .002 }$&$.524_{\pm .010 }$&$.026_{\pm .001 }$&$.190_{\pm .021 }$ & $.297_{\pm .004 }$&$1.740_{\pm .101 }$\\

     &  &  SampleDiv.  & $.910_{\pm .003 }$&$\textbf{.543}_{\pm .014 }$&$.033_{\pm .001 }$&$\textbf{.185}_{\pm .005 }$ & $.298_{\pm .008 }$&$\textbf{1.654}_{\pm .059 }$\\

    \midrule
    
    \multirow{3}{0.12\textwidth}{TreeNet} &  & ind. & $\textbf{.919}_{\pm .002}$  & $.523_{\pm .01}$ &  $.035_{\pm .001}$ & $.234_{\pm .010}$ &  $\textbf{.286}_{\pm .006}$ & $1.990_{\pm .076}$ \\
    
     & 5 & ADP & $.917_{\pm .002}$  & $.535_{\pm .019}$ &  $\textbf{.024}_{\pm .000}$ & $\textbf{.180}_{\pm .031}$ &  $.298_{\pm .005}$ & $1.699_{\pm .132}$ \\
    
     &  & SampleDiv. & $.916_{\pm .002 }$&$\textbf{.545}_{\pm .007 }$&$.030_{\pm .001 }$&$.213_{\pm .012 }$&$.290_{\pm .005 }$&$\textbf{1.659}_{\pm .062 }$\\

    \bottomrule
    
  \end{tabular}
\end{table}

\begin{table}[htbp]
  \caption[Experiments with different Batch Ensemble and Deep Ensemble ensemble sizes on CIFAR-10]{Experiments with different Batch Ensemble and Deep Ensemble ensemble sizes on CIFAR-10. Comparison of the \textit{Sample Diversity} and  \textit{ADP} regularizer with independent training on different architectures under dataset shift.}
  \label{tab:DiffEnsSizeBatchEns}
  \centering

  \begin{tabular}{m{0.11 \textwidth}m{0.035\textwidth}m{0.13\textwidth}m{0.08\textwidth}m{0.08\textwidth}m{0.08\textwidth}m{0.08\textwidth}m{0.08\textwidth}m{0.08\textwidth}}

    \toprule
    Model & Size & Method & \multicolumn{2}{c}{Accuracy $\uparrow$} & \multicolumn{2}{c}{ECE  $\downarrow$} & \multicolumn{2}{c}{NLL $\downarrow$} \\
    && & org. & corr. &  org. & corr. & org. & corr. \\

    \midrule

    \midrule
    
    &  & ind. & $\textbf{.898}_{\pm .005 }$&$.499_{\pm .010 }$&$.046_{\pm .016 }$&$.220_{\pm .048 }$ & $.348_{\pm .027 }$&$1.883_{\pm .109 }$\\

    BatchEns. & 2 & ADP & $.893_{\pm .001 }$&$.518_{\pm .007 }$&$\textbf{.032}_{\pm .004 }$&$\textbf{.118}_{\pm .002 }$ & $.367_{\pm .007 }$&$\textbf{1.596}_{\pm .035 }$\\

    &  & SampleDiv. & $.897_{\pm .001 }$&$\textbf{.533}_{\pm .006 }$&$.038_{\pm .005 }$&$.186_{\pm .008 }$ & $\textbf{.337}_{\pm .011 }$&$1.711_{\pm .036 }$\\

    \midrule
    
    &  & ind. & $.905_{\pm .002 }$&$.503_{\pm .013 }$&$.085_{\pm .015 }$&$.270_{\pm .019 }$ & $.430_{\pm .038 }$&$2.159_{\pm .112 }$\\

    BatchEns. & 3 & ADP & $\textbf{.906}_{\pm .002 }$&$.516_{\pm .010 }$&$\textbf{.035}_{\pm .003 }$&$\textbf{.135}_{\pm .015 }$ & $\textbf{.335}_{\pm .005 }$&$\textbf{1.678}_{\pm .060 }$\\

    &  & SampleDiv. & $.897_{\pm .001 }$&$\textbf{.534}_{\pm .002 }$&$.039_{\pm .018 }$&$.187_{\pm .028 }$ & $.344_{\pm .027 }$&$1.754_{\pm .084 }$\\

    \midrule
    
    &  & ind. & $\textbf{.906}_{\pm .001 }$&$.491_{\pm .001 }$&$.067_{\pm .018 }$&$.261_{\pm .038 }$ & $.377_{\pm .054 }$&$2.117_{\pm .170 }$\\
    
    BatchEns. & 4 & ADP & $.905_{\pm .001 }$&$.513_{\pm .002 }$&$\textbf{.040}_{\pm .011 }$&$\textbf{.157}_{\pm .055 }$ & $\textbf{.360}_{\pm .018 }$&$1.716_{\pm .085 }$\\
    
    &  & SampleDiv. & $.896_{\pm .002 }$&$\textbf{.540}_{\pm .011 }$&$.052_{\pm .010 }$&$.203_{\pm .016 }$ & $.363_{\pm 0.020 }$&$\textbf{1.712}_{\pm .096 }$\\

    \midrule
    
    &  & ind. & $.905_{\pm .001 }$&$.512_{\pm .019 }$&$.097_{\pm .002 }$&$.285_{\pm .014 }$ & $.455_{\pm .003 }$&$2.254_{\pm .099 }$\\

    BatchEns. & 5 & ADP & $\textbf{.906}_{\pm .002 }$&$.517_{\pm .011 }$&$\textbf{.032}_{\pm .008 }$&$\textbf{.171}_{\pm .049 }$ & $.363_{\pm .036 }$&$1.735_{\pm .160 }$\\

    &  & SampleDiv. & $.904_{\pm .000 }$&$\textbf{.545}_{\pm .007 }$&$.037_{\pm .015 }$&$.175_{\pm .032 }$&$\textbf{.343}_{\pm .014 }$&$\textbf{1.649}_{\pm .121 }$\\

    \midrule
    
    &  & ind.  & $.921_{\pm .004 }$&$.519_{\pm .009 }$&$\textbf{.022}_{\pm .001 }$&$.191_{\pm .009 }$&$\textbf{.248}_{\pm .010 }$&$1.786_{\pm .039 }$\\

    DeepEns. & 2 & ADP & $\textbf{.922}_{\pm .002 }$&$.541_{\pm .002 }$&$.027_{\pm .001 }$&$\textbf{.143}_{\pm .007 }$&$.282_{\pm .002 }$&$\textbf{1.599}_{\pm .010 }$\\

    &  & SampleDiv.  & $.921_{\pm .002 }$&$\textbf{.569}_{\pm .012 }$&$.029_{\pm .001 }$&$.158_{\pm .009 }$&$.263_{\pm .005 }$&$1.579_{\pm .062 }$\\

    \midrule

    &  & ind.  & $\textbf{.929}_{\pm .002 }$&$.532_{\pm .009 }$&$\textbf{.021}_{\pm .001 }$&$.174_{\pm .017 }$&$\textbf{.225}_{\pm .005 }$&$1.747_{\pm .058 }$\\

     DeepEns. & 3 & ADP  & $.928_{\pm .001 }$&$.544_{\pm .002 }$&$.029_{\pm .003 }$&$\textbf{.132}_{\pm .008 }$&$.259_{\pm .010 }$&$1.592_{\pm .035 }$\\

    &  & SampleDiv.  & $.926_{\pm .001 }$&$\textbf{.566}_{\pm .014 }$&$.026_{\pm .002 }$&$.149_{\pm .021 }$&$.238_{\pm .002 }$&$\textbf{1.562}_{\pm .066 }$\\

    \midrule

    &  & ind.  & $\textbf{.932}_{\pm .004 }$&$.539_{\pm .005 }$&$\textbf{.023}_{\pm .001 }$&$.157_{\pm .006 }$&$\textbf{.216}_{\pm .009 }$&$1.675_{\pm .038 }$\\

    DeepEns. & 4 &  ADP  & $.929_{\pm .001 }$&$.554_{\pm .002 }$&$.031_{\pm .002 }$&$\textbf{.113}_{\pm .009 }$&$.250_{\pm .008 }$&$1.532_{\pm .006 }$\\

    &  &  SampleDiv.   & $.930_{\pm .002 }$&$\textbf{.575}_{\pm .003 }$&$\textbf{.023}_{\pm .001 }$&$.129_{\pm .007 }$&$.223_{\pm .002 }$&$\textbf{1.499}_{\pm .010 }$\\

    \midrule
    
    &  & ind.  & $\textbf{.936}_{\pm .001 }$&$.543_{\pm .010 }$&$.023_{\pm .001 }$&$.170_{\pm .014 }$&$\textbf{.210}_{\pm .004 }$&$1.783_{\pm .051 }$\\

    DeepEns. & 5 & ADP  & $.933_{\pm .000 }$&$.549_{\pm .005 }$&$.032_{\pm .002 }$&$\textbf{.126}_{\pm .010 }$&$.241_{\pm .004 }$&$1.606_{\pm .037 }$\\

    &  & SampleDiv. & $.933_{\pm .001 }$&$\textbf{.579}_{\pm .004 }$&$\textbf{.022}_{\pm .001 }$&$.134_{\pm .007 }$&$.217_{\pm .008 }$&$\textbf{1.494}_{\pm .026 }$\\
 
    \midrule

    &  & ind.   & $\textbf{.939}_{\pm .001 }$&$.544_{\pm .005 }$&$\textbf{.023}_{\pm .001 }$&$.155_{\pm .003 }$&$\textbf{.191}_{\pm .001 }$&$1.667_{\pm .043 }$\\

    DeepEns. & 11 & ADP$_{\chi^2}$  & $.935_{\pm .004 }$&$.552_{\pm .007 }$&$.028_{\pm .001 }$&$.138_{\pm .007 }$&$.206_{\pm .010 }$&$1.524_{\pm .030 }$\\

    &  & SampleDiv.$_{\chi^2}$  & $.932_{\pm 0.003 }$&$\textbf{.589}_{\pm .003 }$&$.027_{\pm .001 }$&$\textbf{.114}_{\pm .009 }$&$.208_{\pm .009 }$&$\textbf{1.420}_{\pm .034 }$\\

    \bottomrule
    
  \end{tabular}
\end{table}

\end{document}